\documentclass[10pt,logo,copyright]{nvidiatechreport}
\usepackage[authoryear,round]{natbib}

\usepackage[utf8]{inputenc} 
\usepackage[T1]{fontenc}    

\usepackage{parskip}        
\usepackage{url}            
\usepackage{booktabs}       
\usepackage{amsfonts}       
\usepackage{nicefrac}       
\usepackage{microtype}      
\usepackage{xcolor}         
\usepackage[dvipsnames]{xcolor} 
\usepackage{graphicx}
\usepackage{animate}        
\usepackage{subcaption}
\usepackage{tabularx}
\usepackage{makecell}
\usepackage{adjustbox}
\usepackage{setspace}
\newcolumntype{M}[1]{>{\centering\arraybackslash}m{#1}}
\usepackage{float}
\usepackage{tikz}
\usetikzlibrary{positioning,shapes,arrows}
\usepackage{amsmath,amsfonts,bm, bbm,leftindex}
\usepackage{multirow}
\usepackage{comment}
\usepackage{gensymb}
\usepackage{lipsum}
\usetikzlibrary{arrows.meta, positioning, fit}
\usepackage[para]{threeparttable}
\usepackage{tikz}
\usetikzlibrary{tikzmark}

\usepackage[most]{tcolorbox}
\usepackage{fancyvrb}
\usepackage{fvextra}
\usepackage{dashrule}

\usepackage{pifont}
\usepackage{wrapfig}

\usepackage[nameinlink]{cleveref}
\crefname{equation}{Eq.}{Eqs.}
\crefname{figure}{Fig.}{Figs.}
\crefname{section}{Sec.}{Sec.}
\crefname{appendix}{App.}{App.}
\crefname{table}{Tab.}{Tabs.}
\crefname{algorithm}{Algo}{Algo}
\crefname{thm}{Thm}{Thm}
\Crefname{thm}{Thm}{Thm}
\crefname{prop}{Prop}{Prop}
\usepackage[ruled,vlined]{algorithm2e}

\definecolor{darkred}{rgb}{0.7, 0.0, 0.0}

\def\ours{Fast-ThinkAct}

\newcommand{\crefnames}[3]{%
  \@for\next:=#1\do{%
    \expandafter\crefname\expandafter{\next}{#2}{#3}%
  }%
}

\title{Fast-ThinkAct: Efficient Vision-Language-Action Reasoning via Verbalizable Latent Planning}

\author{
    {Chi-Pin Huang}$^{1}$,\; {Yunze Man}$^{2}$,\; {Zhiding Yu},\; {Min-Hung Chen},\; {Jan Kautz},\; {Yu-Chiang Frank Wang}$^{1}$,\; {Fu-En Yang}\\
    \normalsize NVIDIA
}

\correspondingauthor{X}

\begin{document}

\maketitle

\vspace{-5mm}

\begin{abstract}
Vision-Language-Action (VLA) tasks require reasoning over complex visual scenes and executing adaptive actions in dynamic environments. While recent studies on reasoning VLAs show that explicit chain-of-thought (CoT) can improve generalization, they suffer from high inference latency due to lengthy reasoning traces. We propose Fast-ThinkAct, an efficient reasoning framework that achieves compact yet performant planning through verbalizable latent reasoning. Fast-ThinkAct learns to reason efficiently with latent CoTs by distilling from a teacher, driven by a preference-guided objective to align manipulation trajectories that transfers both linguistic and visual planning capabilities for embodied control. This enables reasoning-enhanced policy learning that effectively connects compact reasoning to action execution. Extensive experiments across diverse embodied manipulation and reasoning benchmarks demonstrate that Fast-ThinkAct achieves strong performance with up to 89.3\% reduced inference latency over state-of-the-art reasoning VLAs, while maintaining effective long-horizon planning, few-shot adaptation, and failure recovery.\\

\textbf{Links:} \hspace{2pt}
{
\hypersetup{urlcolor=nvidiagreen}
\href{https://jasper0314-huang.github.io/fast-thinkact/}{Project Page}
}

\end{abstract}
\abscontent
\section{Introduction}
\label{sec:intro}

\begin{wrapfigure}{R}{0.55\textwidth}
    \vspace{-3mm}

    \centering
    \includegraphics[width=1.0\linewidth]{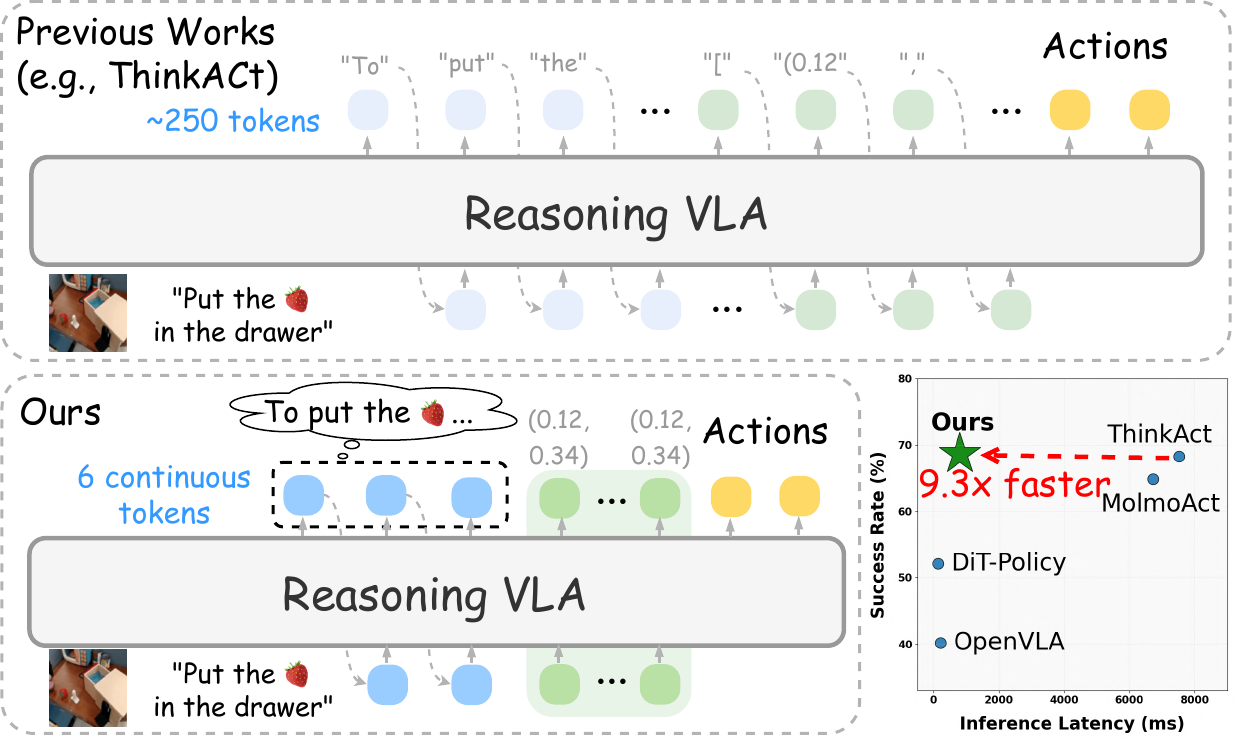}

    \vspace{1.5mm}
    
    \caption{\textbf{Overview of \ours{}.} Previous reasoning VLAs generate lengthy reasoning traces ($\sim$250 tokens). Our approach learns compact continuous tokens (e.g., 6) (\textcolor{cyan}{blue}) and parallel spatial tokens (\textcolor{YellowGreen}{green}) as internal reasoning. The bottom-right plot shows that we achieve $9.3\times$ faster inference than ThinkAct-7B~\cite{huang2025thinkact}, while delivering improved performance on the SimplerEnv-Google benchmark.}
    \label{fig:teaser}
\end{wrapfigure}

Recent large vision-language models (VLMs)~\cite{liu2023visual,comanici2025gemini,liu2024nvila,bai2025qwen2,shi2024eagle,li2025eagle,chen2025eagle,wang2025internvl3,xie2024show} have achieved remarkable capabilities in visual-language understanding across diverse multimodal tasks. To extend these capabilities to embodied-centric tasks, recent works leverage large-scale robot demonstrations~\cite{o2024open} to develop Vision-Language-Action (VLA) foundation models~\cite{brohan2022rt,brohan2023rt,team2024octo,bjorck2025gr00t,li2025hamster,black2024pi_0,yang2025magma,kim24openvla}. These VLA tasks require agents to perceive complex visual scenes, reason over spatial and temporal contexts, and execute adaptive actions within dynamic environments, demanding robust long-horizon planning and contextual adaptation. However, as these VLA models primarily rely on supervised training from action data, they excel at basic skills (e.g., pick-and-place) but struggle to generalize beyond training distributions, such as long-horizon planning, self-correction from failures, and adaptation to novel scenarios, due to the impracticality of collecting exhaustive robot demonstrations.

Reasoning VLAs~\cite{zawalski2024robotic,zhao2025cot,lee2025molmoact,wu2025you,qu2025eo,huang2025thinkact,kim2025robot} address these limitations by incorporating intermediate thinking processes, improving generalization and task-solving capability. Supervised chain-of-thought (CoT) methods~\cite{zawalski2024robotic,zhao2025cot,lee2025molmoact,qu2025eo} address this by learning from intermediate reasoning annotations. These approaches can be categorized into textual reasoning methods that leverage off-the-shelf LLMs and VLMs to generate pseudo CoT labels~\cite{zawalski2024robotic}, and visual reasoning methods that generate structured visual reasoning representations such as sub-goal images, image depth, and 2D visual traces~\cite{zhao2025cot,lee2025molmoact}. However, these supervised approaches require substantial reasoning annotations and remain limited by training data coverage. To address this, ThinkAct~\cite{huang2025thinkact} employs RL-based reasoning~\cite{shao2024deepseekmath} to generate long textual CoTs guided by action-aligned visual rewards. While these reasoning methods effectively improve task generalization and planning capabilities, they require generating lengthy chain-of-thought steps that introduce substantial reasoning latency, which hampers embodied applications with \emph{real-time} requirements.

In embodied AI applications such as robotic manipulation and autonomous driving, agents must make rapid decisions at high frequencies (e.g., 1-15 Hz)~\cite{guan2025efficient}. However, generating lengthy reasoning traces can take several seconds per decision (e.g., 0.1 Hz)~\cite{huang2025thinkact, lee2025molmoact}, creating a critical bottleneck that limits real-time performance~\cite{guan2025efficient, yu2025survey} and poses safety risks in time-critical scenarios~\cite{wang2025alpamayo}. To mitigate this efficiency bottleneck while preserving reasoning capabilities, very recent works~\cite{chen2025training, yu2025survey, guan2025efficient} have explored approaches to reduce inference latency in embodied reasoning. For instance, ECoT-Lite~\cite{chen2025training} proposes reasoning dropout to accelerate inference, yet directly reducing textual reasoning length risks performance degradation due to critical information loss. How to preserve reasoning capability while enabling compact representations that properly capture essential spatial-temporal dynamics remains a crucial challenge for reasoning VLA models.

In this paper, we propose \textit{\ours{}}, an efficient embodied reasoning framework for Vision-Language-Action tasks that achieves compact yet expressive planning through verbalizable latent reasoning. As depicted in Figure~\ref{fig:teaser}, unlike prior reasoning VLAs that generate lengthy explicit textual CoT traces, we introduce reward-guided preference distillation with visual trajectory alignment to compress linguistic and visual planning into compact continuous latents that enable implicit internal reasoning. Our student VLM encodes reasoning into compact latents decodable by a verbalizer, enabling preference-based optimization that leverages RL-derived reward signals to distill high-quality reasoning patterns from a textual teacher VLM while suppressing low-quality ones. We further align trajectory latents between teacher and student to transfer visual planning capabilities essential for embodied control. Once trained, the student VLM enables reasoning-enhanced policy learning that bridges implicit multimodal planning with action execution, achieving significantly faster inference while outperforming existing reasoning VLAs.

Our contributions can be summarized as follows:
\begin{itemize}
    \item We propose \textit{\ours{}}, an efficient reasoning framework that compresses reasoning into verbalizable latent thoughts while maintaining expressive planning abilities.
    
    \item We introduce preference-guided distillation with manipulation trajectory alignment that compresses linguistic and visual planning into compact continuous latents.
    
    \item We bridge high-level visual planning to low-level action execution through reasoning-enhanced policy learning guided by manipulation trajectory latents.
    
    \item We achieve up to 89.3\% inference latency reduction over state-of-the-art reasoning VLAs while maintaining strong performance across diverse embodied benchmarks.
\end{itemize}
\section{Related Works}
\label{sec:related_works}

\subsection{Vision-Language-Action (VLA) Models}

\noindent\textbf{Foundation VLAs.}
Vision-Language-Action (VLA) models~\cite{brohan2022rt,brohan2023rt,team2024octo,bjorck2025gr00t,li2025hamster,black2024pi_0,yang2025magma,pertsch2025fast,driess2025knowledge,bu2025agibot,team2025gemini,wang2025vla} have recently emerged as a promising paradigm for embodied AI by training vision-language backbones on large-scale robot demonstrations. Works such as OpenVLA~\cite{kim24openvla} and $\pi_0$~\cite{black2024pi_0} achieve language-conditioned manipulation through end-to-end policy learning, while Magma~\cite{yang2025magma} co-trains on heterogeneous human and robot data. HAMSTER~\cite{li2025hamster} and TraceVLA~\cite{zheng2024tracevla} further leverage 2D visual trajectories to boost spatial-action connections. Despite success on routine manipulation, these imitation-based approaches struggle with long-horizon planning and generalization to novel scenarios due to limited training data coverage.

\vspace{2mm}

\noindent\textbf{Reasoning VLAs.}
To overcome these limitations, recent works~\cite{zawalski2024robotic,zhao2025cot,lee2025molmoact,wu2025you,qu2025eo,huang2025thinkact,kim2025robot,yuan2025embodied,abdolmaleki2025gemini} integrate explicit reasoning mechanisms into VLA architectures. Supervised approaches~\cite{zawalski2024robotic,zhao2025cot,lee2025molmoact,qu2025eo} introduce intermediate reasoning through chain-of-thought annotations. Embodied CoT~\cite{zawalski2024robotic} and Hi-Robot~\cite{shi2025hi} synthesize reasoning labels via pretrained foundation models. To perform vision-centric reasoning~\cite{man2025argus,sarch2025grounded} beyond pure text, CoT-VLA~\cite{zhao2025cot} employs visual goal generation and MolmoAct~\cite{lee2025molmoact} structures reasoning by spatial representations. Additionally, EO-1~\cite{qu2025eo} introduces interleaved vision-language-action pre-training to bridge reasoning and interaction. Recent works~\cite{yuan2025embodied,huang2025thinkact} alternatively leverage reinforcement fine-tuning to generate reasoning chains with designed rewards. Despite improved generalization, these reasoning VLAs suffer from high inference latency and inevitably introduce extraneous information that degrades action quality.

\subsection{Efficient Reasoning}

To address the inference latency of reasoning, recent LLM research explores various efficiency techniques~\cite{lee2025vlsi,dai2025stable,yuan2025efficient,xiang2025just,aggarwal2025l1,lee2025unified}. For example, RL-based approaches~\cite{dai2025stable,yuan2025efficient,xiang2025just,aggarwal2025l1} introduce length penalties to encourage shorter reasoning chains, though such methods can suffer from training instability. Beyond length control, latent reasoning methods~\cite{hao2024training,shen2025codi,zhang2025soft,cheng2024compressed,xu2025softcot} enable reasoning in continuous spaces, such as Coconut~\cite{hao2024training} using hidden states as continuous thoughts, CODI~\cite{shen2025codi} distilling explicit CoT into continuous space via teacher-student alignment, and Soft Thinking~\cite{zhang2025soft} generating weighted concept tokens. However, these LLM techniques cannot directly transfer to VLA tasks due to the need for spatial-temporal understanding and bridging semantic reasoning with embodied control.
Recently, ECoT-Lite~\cite{chen2025training} proposes reasoning dropout to accelerate embodied reasoning by skipping test-time reasoning traces. However, reasoning dropout can lead to inconsistent planning as it builds on supervised embodied CoT. Our proposed \ours{} distills reasoning into compact latent representations that naturally encode multimodal information, enabling robust reasoning-enhanced policy learning.
\section{Method}
\label{sec:method}


\begin{figure*}[t!]
    \centering
    \includegraphics[width=1.0\linewidth]{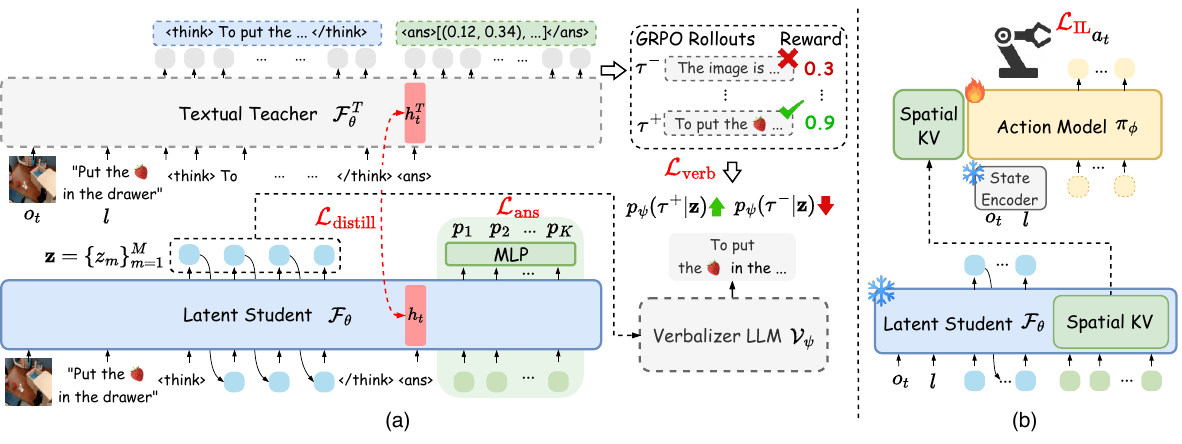}
    \caption{\textbf{Overview of Fast-ThinkAct.} (a) Given observation $o_t$ and instruction $l$, the Textual Teacher VLM $\mathcal{F}_\theta^T$ generates explicit reasoning chains. The Latent Student VLM $\mathcal{F}_\theta$ distills these into compact latent tokens $\mathbf{z}$ guided by reward preferences. Verbalizer LLM $\mathcal{V}_\psi$ decodes latents to text for preference-based learning via $\mathcal{L}_{\text{verb}}$, while $\mathcal{L}_{\text{distill}}$ transfers visual planning capability from teacher, and spatial tokens enable parallel visual trajectory prediction via $\mathcal{L}_{\text{ans}}$, ensuring latents are verbalizable and grounded in visual planning. (b) Reasoning-Enhanced Policy Learning. The Action Model $\pi_\phi$ is trained with $\mathcal{L}_{\text{IL}}$ while freezing the latent student $\mathcal{F}_\theta$ and state encoder.}
    \label{fig:method}
\end{figure*}

\subsection{Problem Formulation}

We first define the setting and notations. At each timestep $t$, given a language instruction $l$, the model observes a visual input $o_t$ and generates an action chunk $a_t$, represented as a sequence of continuous robot control vectors (e.g., 7- or 14-DOF for single- or bimanual robots, respectively).

To address this problem, we propose \ours{}, an efficient reasoning framework that bridges high-level planning with low-level action execution. Our approach employs a VLM $\mathcal{F}_\theta$ to perform reasoning in \textit{continuous latent space}, integrated with an action model $\pi_\phi$ for executable action generation. Specifically, $\mathcal{F}_\theta$ processes observation-instruction pairs $(o_t, l)$ through latent chain-of-thought (CoT) reasoning to produce a compact visual plan latent $c_t$ that encapsulates the intended trajectory in visual space (Sec.~\ref{ssec:reasoning}). This $c_t$ subsequently guides $\pi_\phi$ to predict executable actions $a_t$ (Sec.~\ref{ssec:action}). By distilling reasoning into a continuous latent space rather than discrete text, \ours{} achieves significantly improved inference efficiency while enhancing action performance through better preservation of spatial and visual information.

\subsection{Efficient Embodied Reasoning}\label{ssec:reasoning}

To enable efficient embodied reasoning that meets the real-time requirements of embodied AI tasks, we aim to compress long textual CoTs into a compact set of continuous latent representations. However, compressing reasoning traces into latents is challenging, as there is no direct supervision signal in the latent space to guide what reasoning patterns should be encoded.

\subsubsection{Verbalizable Latent CoT by Reward Preferences}\label{sssec:verbalize}

To address this challenge, we propose to perform distillation in natural language space by introducing a verbalizer LLM that decodes latents into verbalizable reasoning. This approach grounds latent learning in an interpretable textual form, ensuring that the learned latents faithfully preserve the underlying reasoning structure. Since reasoning traces generated by the teacher model $\mathcal{F}_\theta^T$ exhibit varying quality, we adopt a preference-based learning framework that exploits reward signals from the teacher's GRPO training to guide the latent student $\mathcal{F}_\theta$ toward high-quality reasoning patterns while suppressing low-quality ones.

Specifically, we employ a teacher-student framework where a textual teacher model $\mathcal{F}_\theta^T$ first learns explicit reasoning through GRPO~\cite{shao2024deepseekmath} training by maximizing:
\begin{equation}
    \mathcal{J}_\text{GRPO}(\theta) = \mathbb{E}_{\tau\sim\mathcal{F}^T_\theta}\Big[ \min \big( r_\theta(\tau)A(\tau), \text{clip}(r_\theta(\tau), 1-\epsilon, 1+\epsilon )A(\tau) \big) \Big],
\end{equation}
where $\tau$ denotes a reasoning trace and $r_\theta(\tau) = \frac{\mathcal{F}^T_\theta(\tau)}{\mathcal{F}^T_{\text{old}}(\tau)}$ is the probability ratio. The advantage function for group rewards $\{ R_i \}_{i\in G(\tau)}$ is represented as:
\begin{equation}
     A(\tau)=\frac{R_\tau - \text{mean}(\{ R_i \}_{i\in G(\tau)})}{\text{std}(\{ R_i \}_{i\in G(\tau)})}.
\end{equation}

This training process produces textual CoTs with varying quality, where the advantage function $A(\tau)$ naturally serves as a quality indicator. To construct preference pairs for distillation, we select the highest and lowest advantage traces from each rollout group:
\begin{align}
\tau^+ = \arg\max_{\tau \in G} A(\tau) \text{ and } \tau^- = \arg\min_{\tau \in G} A(\tau).
\end{align}

Instead of generating textual tokens, the student model $\mathcal{F}_\theta$ performs latent reasoning by autoregressively generating $M$ continuous latent vectors $\mathbf{z} = \{z_m\}_{m=1}^M$ with $z_m \in \mathbb{R}^d$, where $d$ is the hidden size. We then train the verbalizer LLM $\mathcal{V}_\psi$ to decode these latents $\mathbf{z}$ into natural language.  The training objective encourages the verbalizer to assign a higher likelihood to decoding latents into high-quality reasoning $\tau^+$ than low-quality reasoning $\tau^-$. Inspired by DPO~\cite{rafailov2023direct}, we formulate this as an optimization guided by the reward preferences:
\begin{equation}
\mathcal{L}_{\text{verb}}
=
-\mathbb{E} \Big[ \log \sigma\Big(
\beta\big(
\log \tfrac{p_\psi(\tau^{+}\mid \mathbf{z})}
          {p_{\text{ref}}(\tau^{+})} -\log \tfrac{p_\psi(\tau^{-}\mid \mathbf{z})}
            {p_{\text{ref}}(\tau^{-})}
\big)\Big)\Big],
\label{eq:verbalize}
\end{equation}
where $p_{\text{ref}}$ is the reference model (i.e., $\mathcal{V}_\psi$ without latent conditioning), $\sigma$ is the sigmoid function, and $\beta=0.1$ controls preference strength. This encourages the student VLM $\mathcal{F}_\theta$ to encode latents that the verbalizer decodes into high-quality reasoning while suppressing low-quality patterns.

\subsubsection{Action-Aligned Visual Plan Distillation}\label{sssec:distill}
While the verbalizer loss (Eq.~\ref{eq:verbalize}) enables the student $\mathcal{F}_\theta$ to capture high-level reasoning patterns, it does not explicitly ensure that latent representations encode the visual planning capability crucial for embodied control. To address this, we introduce action-aligned visual plan distillation to transfer the teacher $\mathcal{F}_\theta^T$'s spatial reasoning ability to the student $\mathcal{F}_\theta$.

We distill spatial reasoning from the teacher, which is trained with trajectory-level rewards (e.g., goal completion and trajectory alignment~\cite{huang2025thinkact}) for grounded visual planning. We align the trajectory-level representations by minimizing the L2 distance between hidden states of the \texttt{<answer>} token that encodes the visual plan:
\begin{equation}
    \mathcal{L}_{\text{distill}} = \| h_t^T - h_t \|_2^2,
\end{equation}
where $h_t^T$ and $h_t$ are the hidden states from teacher (corresponding to $\tau^+$) and student, respectively.

To enable efficient parallel trajectory prediction, unlike the textual teacher that autoregressively generates verbose text sequences of waypoints $\{p_k\}_{k=1}^K$ with $p_k \in [0, 1]^2$ (tokenized into 60-70 tokens when $K=5$), the student uses $K$ learnable spatial tokens $\{\mathbf{s}_i\}_{i=1}^K$ appended to the reasoning latent sequence, with each output hidden state simultaneously projected to a waypoint via an MLP. The total objective for training $\mathcal{F}_\theta$ combines all three components:
\begin{equation}
\begin{aligned}
    \mathcal{L}_{\text{student}} &= \mathcal{L}_{\text{verb}} + \mathcal{L}_{\text{distill}} + \mathcal{L}_{\text{ans}}, \quad \text{where } \\
     \mathcal{L}_{\text{ans}} &= \sum_{i=1}^K \| p_i - \hat{p}_i \|_2^2, \text{ with } p_i = \text{MLP}(h^\prime(\mathbf{s}_i)),
\end{aligned}
\end{equation}
where $h^\prime(\mathbf{s}_i)$ denotes the output hidden state of the $i$-th spatial token and $\hat{p}_i$ are ground-truth waypoints. Through this unified framework, the student model $\mathcal{F}_\theta$ performs compact yet expressive latent reasoning and generates visual trajectory plans efficiently.

\subsection{Reasoning-Enhanced Policy Learning}\label{ssec:action}
After the student VLM $\mathcal{F}_\theta$ performs compact latent reasoning and generates visual trajectory planning through spatial tokens, we leverage these representations to guide a diffusion Transformer-based action model $\pi_\phi$ (e.g., RDT~\cite{liu2024rdt}) for action prediction. To bridge the high-level visual planning with low-level action generation, we connect the visual latent planning $c_t$ encoded in the key-value cache corresponding to the spatial tokens to the action model.

Specifically, we extract visual latent planning $c_t$ from the KV cache of spatial tokens in earlier VLM layers (since $\mathcal{F}_\theta$ has more layers than $\pi_\phi$) and concatenate with KV pairs from the action model's state encoder. The action model's cross-attention then attends to both the visual planning context and state observations. We post-train on action-annotated robot data by \textit{freezing} $\mathcal{F}_\theta$ and the state encoder while updating only $\pi_\phi$ with the imitation learning objective:
\begin{equation}
\mathcal{L}_{\text{IL}}(\phi) =  \ell\left(\pi_\phi(o_t, l, c_{t}), \hat{a}_t \right),
\end{equation}
where $\ell$ denotes the denoising objective for diffusion policy and $\hat{a}_t$ is the ground-truth action. Through this post-training, the action model effectively translates visual planning from compact latent reasoning into low-level robot actions.

\begin{figure*}[t!]
    \centering
    \includegraphics[width=1.0\linewidth]{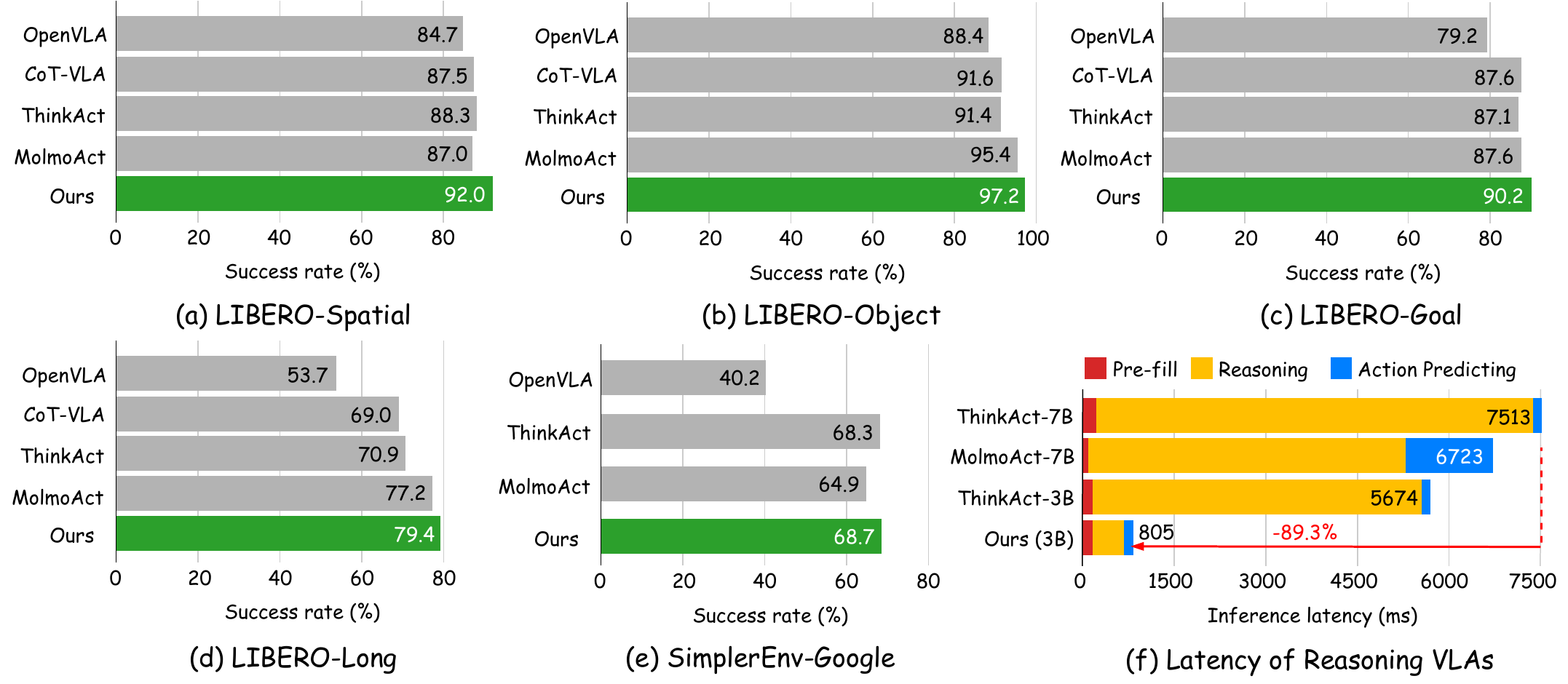}
    \caption{\textbf{Evaluation of robot manipulation and reasoning efficiency.} (a)-(e) Success rates on LIBERO~\cite{liu2023libero} and SimplerEnv~\cite{li24simpler} benchmarks compared with state-of-the-art 7B reasoning VLAs. (f) Latency comparison across 3B and 7B reasoning VLAs. Our approach achieves up to 89.3\% inference latency reduction while maintaining superior task success rates.}
    \label{fig:libero_simpler}
\end{figure*}

\subsection{Learning Strategy and Inference}\label{ssec:inference}

\noindent\textbf{Training Strategy.}
We initialize both teacher $\mathcal{F}_\theta^T$ and student $\mathcal{F}_\theta$ from the same checkpoint obtained through SFT and CoT-SFT on a pre-trained VLM. The teacher is trained with GRPO using action-aligned rewards~\cite{huang2025thinkact}, while the student is trained with $\mathcal{L}_\text{student}$ to compress reasoning into compact latents. We then connect the trained $\mathcal{F}_\theta$ with action model $\pi_\phi$ (initialized from~\cite{liu2024rdt}) by freezing $\mathcal{F}_\theta$ and the state encoder while updating the latent projector and $\pi_\phi$ with $\mathcal{L}_{\text{IL}}$ on large-scale robotic data. For target environment adaptation (e.g., LIBERO~\cite{liu2023libero}, RoboTwin2.0~\cite{chen2025robotwin2}), we fine-tune on environment-specific demonstrations.

\noindent\textbf{Inference.} 
The $\mathcal{F}_\theta$ processes $(o_t, l)$ by compact latent reasoning, generating visual trajectories via $K$ spatial tokens. The visual latent planning $c_t$, extracted from the spatial tokens' KV cache, conditions $\pi_\phi$ to predict actions $a_t$. Inference requires only $\mathcal{F}_\theta$ and $\pi_\phi$; the verbalizer $\mathcal{V}_\psi$ is used solely during training and optionally for interpretability.
\section{Experiment}
\label{sec:experiments}

\subsection{Experimental Setup}

\paragraph{Implementation Details.}

We use Qwen2.5-VL 3B~\cite{bai2025qwen2} as the VLM backbone. The SFT stage runs for 1 epoch with batch size 64 and learning rate $1\text{e}{-5}$, followed by CoT-SFT for 15K iterations with the same hyperparameters. For teacher-student training, both $\mathcal{F}_\theta^T$ and $\mathcal{F}_\theta$ are initialized from the CoT-SFT checkpoint and trained for 4,500 iterations with batch size 128 and learning rate $1\text{e}{-6}$. The teacher is optimized with GRPO~\cite{shao2024deepseekmath} using action-aligned visual rewards~\cite{huang2025thinkact} and QA-style rewards (detailed in supplementary material). For the first 3,000 iterations of the student training, we train the verbalizer $\mathcal{V}_\psi$ with standard language modeling loss, then switch to $\mathcal{L}_{\text{verb}}$ for the remaining 1,500 iterations. For reasoning-enhanced policy learning, we initialize $\pi_\phi$ from DiT-Policy~\cite{chi2023diffusion} pre-trained on OXE~\cite{o2024open} for SimplerEnv, and from RDT~\cite{liu2024rdt} for LIBERO and RoboTwin2.0. A linear projection adapts the VLM's KV cache to the action model dimension (1,024 for DiT-Policy and 2,048 for RDT). Training runs for 20K iterations with batch size 256 and learning rate $1\text{e}{-4}$. All diffusion hyperparameters follow those of the respective action models. All experiments are conducted on 16 NVIDIA A100 GPUs with 80 GB memory.

\paragraph{Training Datasets and Evaluation Benchmarks.}

For reasoning VLM training, we utilize single-arm visual trajectories labeled by~\cite{lee2025molmoact} and dual-arm visual trajectories from the AIST dataset~\cite{aist2025aist}, along with QA tasks from PixMo~\cite{deitke2024molmo}, RoboFAC~\cite{lu2025robofac}, RoboVQA~\cite{sermanet2024robovqa}, ShareRobot~\cite{ji2025robobrain}, EgoPlan~\cite{chen2023egoplan}, and Video-R1~\cite{feng2025video}. For reasoning-enhanced policy learning, we use action data from the OXE dataset~\cite{o2024open} (following OpenVLA~\cite{kim24openvla}) when training with DiT-Policy, and augment with bimanual data from the static Aloha dataset~\cite{shi2023waypoint, zhao2023learning} when training with RDT.

We evaluate \ours{} on four embodied reasoning benchmarks and three robot manipulation benchmarks. For embodied reasoning, we use EgoPlan-Bench2~\cite{qiu2024egoplan2} (accuracy on multiple-choice questions), RoboVQA~\cite{sermanet2024robovqa} (BLEU score~\cite{papineni2002bleu}), OpenEQA~\cite{majumdar2024openeqa}, and RoboFAC~\cite{lu2025robofac} (both using LLM-based scoring). Notably, RoboVQA and RoboFAC contain videos captured from real robots. For robot manipulation, we evaluate on SimplerEnv~\cite{li24simpler}, which demonstrates strong correlation with real-world performance, LIBERO~\cite{liu2023libero} covering diverse manipulation tasks including long-horizon scenarios, and RoboTwin2.0~\cite{chen2025robotwin2} for complex bimanual manipulation. All robot manipulation tasks use task success rate as the metric. Additional details are provided in the supplementary material.

\subsection{Quantitative Evaluation}

\paragraph{Robot Manipulation.}

\begin{table*}[t!]
    \centering
    \caption{\textbf{Quantitative evaluation on RoboTwin2.0~\cite{chen2025robotwin2}.} E and H denote easy and hard settings (without/with domain randomization). Background colors indicate task length based on expert demonstrations: \colorbox{green!20}{short (80-100)}, \colorbox{yellow!20}{medium (110-220)}, \colorbox{red!20}{long (270-470)} steps.}
    \resizebox{\textwidth}{!}{%
        \begin{tabular}{l|
        cc|cc|cc|
        cc|cc|cc|
        cc|cc|cc|cc|cccc}
            \toprule
            \multirow{2}{*}{Model} 
            & \multicolumn{2}{c|}{\cellcolor{green!20}\makecell{click\\alarm}} 
            & \multicolumn{2}{c|}{\cellcolor{green!20}\makecell{click\\bell}} 
            & \multicolumn{2}{c|}{\cellcolor{green!20}\makecell{turn\\switch}} 
            & \multicolumn{2}{c|}{\cellcolor{yellow!20}\makecell{adjust\\bottle}} 
            & \multicolumn{2}{c|}{\cellcolor{yellow!20}\makecell{beat\\block}} 
            & \multicolumn{2}{c|}{\cellcolor{yellow!20}\makecell{handover\\mic}} 
            & \multicolumn{2}{c|}{\cellcolor{red!20}\makecell{handover\\block}} 
            & \multicolumn{2}{c|}{\cellcolor{red!20}\makecell{hanging\\mug}} 
            & \multicolumn{2}{c|}{\cellcolor{red!20}\makecell{stack\\blocks~two}} 
            & \multicolumn{2}{c|}{\cellcolor{red!20}\makecell{stack\\bowls~three}} 
            & \multicolumn{2}{c}{Average} \\
            \cmidrule{2-23}
            & \cellcolor{green!20}E & \cellcolor{green!20}H 
            & \cellcolor{green!20}E & \cellcolor{green!20}H 
            & \cellcolor{green!20}E & \cellcolor{green!20}H 
            & \cellcolor{yellow!20}E & \cellcolor{yellow!20}H 
            & \cellcolor{yellow!20}E & \cellcolor{yellow!20}H 
            & \cellcolor{yellow!20}E & \cellcolor{yellow!20}H 
            & \cellcolor{red!20}E & \cellcolor{red!20}H 
            & \cellcolor{red!20}E & \cellcolor{red!20}H 
            & \cellcolor{red!20}E & \cellcolor{red!20}H 
            & \cellcolor{red!20}E & \cellcolor{red!20}H 
            & E & H \\
            \midrule
            DP~\cite{chi2023diffusion} & 61 & 5 & 54 & 0 & 36 & 1 & 97 & 0 & 42 & 0 & 53 & 0 & 10 & 0 & 8 & 0 & 7 & 0 & 63 & 0 & 43.1 & 0.6 \\
            ACT~\cite{zhao2023learning} & 32 & 4 & 58 & 3 & 5 & 2 & 97 & 23 & 56 & 3 & 85 & 0 & 42 & 0 & 7 & 0 & 25 & 0 & 48 & 0 & 45.5 & 3.5 \\
            $\pi_0$~\cite{black2024pi_0} & 63 & 11 & 44 & 3 & 27 & 23 & 90 & 56 & 43 & 21 & 98 & 13 & 45 & 8 & 11 & 3 & 42 & 1 & 66 & 24 & 52.9 & 16.3 \\
            RDT~\cite{liu2024rdt} & 61 & 12 & 80 & 9 & 35 & 15 & 81 & 75 & 77 & 37 & 90 & 31 & 45 & 14 & 23 & 16 & 21 & 2 & 51 & 17 & 56.4 & 22.8 \\
            ThinkAct~\cite{huang2025thinkact} & 64 & 13 & 84 & 11 & 40 & 19 & 94 & 70 & 79 & 33 & 92 & 40 & 56 & 15 & 31 & 18 & 30 & 5 & 54 & 23 & 62.4 & 24.7 \\
            \midrule
            \textbf{\ours{}} & 70 & 17 & 82 & 12 & 37 & 21 & 92 & 72 & 82 & 33 & 99 & 42 & 65 & 15 & 30 & 22 & 45 & 5 & 55 & 25 & \textbf{65.7} & \textbf{26.4} \\
            \bottomrule
        \end{tabular}%
    }
    \label{tab:robotwin}
\end{table*}
\begin{table*}[t]
\centering
\caption{\textbf{Quantitative evaluation on EgoPlan-Bench2~\cite{qiu2024egoplan2}, RoboVQA~\cite{sermanet2024robovqa}, and OpenEQA~\cite{majumdar2024openeqa} benchmarks for embodied reasoning.}}
\resizebox{1.0\textwidth}{!}{
\small
\begin{tabular}{lccccccccccccc}
\toprule
\textbf{Method} & \multicolumn{5}{c}{\textbf{EgoPlan-Bench2}} & \multicolumn{5}{c}{\textbf{RoboVQA}} & \textbf{OpenEQA} & \cellcolor{gray!15}\textbf{Overall} \\
\cmidrule(lr){2-6} \cmidrule(lr){7-11}
 & Daily. & Work. & Rec. & Hobbies & Avg. & B-1 & B-2 & B-3 & B-4 & B-Avg. & Score & \cellcolor{gray!15}Avg. \\
\midrule
GPT-4V~\cite{achiam2023gpt} & 36.7 & 27.7 & 33.9 & 32.5 & 32.6 & 32.2 & 26.5 & 24.7 & 23.9 & 26.8 & 49.6 & \cellcolor{gray!15}36.4 \\
Gemini-2.5-Flash~\cite{comanici2025gemini} & 44.2 & 42.3 & 43.2 & 39.1 & 42.4 & 39.1 & 31.6 & 22.9 & 22.1 & 28.9 & 45.3 & \cellcolor{gray!15}38.9 \\
\midrule
InternVL2.5-2B~\cite{chen2024expanding} & 30.9 & 27.8 & 28.6 & 33.1 & 30.1 & 36.6 & 33.7 & 31.0 & 29.4 & 32.7 & 47.1 & \cellcolor{gray!15}36.6 \\
InternVL3-2B~\cite{zhu2025internvl3} & 36.9 & 29.9 & 35.6 & 31.5 & 33.4 & 34.4 & 33.9 & 33.5 & 33.3 & 33.8 & 48.8 & \cellcolor{gray!15}38.7 \\
NVILA-2B~\cite{liu2024nvila} & 34.6 & 26.7 & 33.3 & 31.6 & 31.4 & 38.7 & 34.3 & 31.1 & 29.2 & 33.3 & 47.0 & \cellcolor{gray!15}37.2 \\
Qwen2.5-VL-3B~\cite{bai2025qwen2} & 29.0 & 27.0 & 30.2 & 28.9 & 28.5 & 42.5 & 36.3 & 28.7 & 31.8 & 34.8 & 43.4 & \cellcolor{gray!15}35.6 \\
Magma-8B~\cite{yang2025magma} & 32.1 & 25.7 & 34.4 & 29.3 & 29.8 & 38.6 & 31.5 & 28.1 & 26.7 & 31.2 & 49.1 & \cellcolor{gray!15}36.7 \\
RoboBrain2.0-3B~\cite{team2025robobrain2} & 45.3 & 37.6 & 45.9 & 39.7 & 41.8 & 54.4 & 47.7 & 43.1 & 41.0 & 46.5 & 50.1 & \cellcolor{gray!15}46.1 \\
ThinkAct-3B~\cite{huang2025thinkact} & 46.6 & 41.4 & 45.9 & 42.5 & 44.0 & 62.4 & 57.3 & 52.0 & 49.6 & 55.3 & 48.9 & \cellcolor{gray!15}49.4 \\
\midrule
\textbf{Fast-ThinkAct-3B} & \textbf{50.3} & \textbf{44.3} & \textbf{46.4} & \textbf{43.2} & \textbf{46.4} & \textbf{70.1} & \textbf{63.0} & \textbf{57.2} & \textbf{53.0} & \textbf{60.8} & \textbf{51.2} & \cellcolor{gray!15}\textbf{52.8} \\
\bottomrule
\end{tabular}
}
\label{tab:egoplan_robo_results}
\vspace{-2mm}
\end{table*}

We evaluate \ours{} on robotic manipulation using LIBERO~\cite{liu2023libero} and SimplerEnv~\cite{li24simpler} benchmarks. LIBERO covers diverse subtasks, including Spatial, Object, Goal, and Long, while SimplerEnv provides a simulated benchmark with strong real-world correlation, featuring variations in lighting, object appearance, and camera viewpoints. As shown in Fig.~\ref{fig:libero_simpler}(a)-(e), \ours{} consistently outperforms all baselines, achieving the highest success rates across all LIBERO subtasks and SimplerEnv-Google. This includes substantial improvements over foundation VLAs such as OpenVLA~\cite{kim24openvla}, and reasoning VLAs including CoT-VLA~\cite{zhao2025cot}, ThinkAct~\cite{huang2025thinkact}, and MolmoAct~\cite{lee2025molmoact}. Moreover, as shown in Fig.~\ref{fig:libero_simpler}(f), our compact latent reasoning achieves 89.3\% and 88.0\% latency reduction compared to ThinkAct-7B~\cite{huang2025thinkact} and MolmoAct-7B~\cite{lee2025molmoact} respectively, and $7\times$ faster inference than ThinkAct-3B, demonstrating substantial efficiency gains without sacrificing performance.

To further validate \ours{} on more complex scenarios, we evaluate on RoboTwin2.0~\cite{chen2025robotwin2}, a challenging bimanual manipulation benchmark requiring long-horizon planning. As shown in Tab.~\ref{tab:robotwin}, \ours{} significantly outperforms previous VLAs including DP~\cite{chi2023diffusion}, ACT~\cite{zhao2023learning}, $\pi_0$~\cite{black2024pi_0}, RDT~\cite{liu2024rdt}, and ThinkAct~\cite{huang2025thinkact} across both easy and hard settings. Compared to RDT, \ours{} achieves 9.3\% and 3.6\% higher success rates on easy and hard settings, respectively. Against the reasoning VLA ThinkAct, it improves success by 3.3\% and 1.7\% while maintaining substantially higher efficiency, as shown in Fig.~\ref{fig:libero_simpler}(f). These results demonstrate that our compact reasoning design enables both superior accuracy and computational efficiency on complex bimanual manipulation tasks.

\paragraph{Embodied Reasoning.}

In Tab.~\ref{tab:egoplan_robo_results}, we evaluate the reasoning capabilities of \ours{} in embodied scenarios across three benchmarks: EgoPlan-Bench2~\cite{qiu2024egoplan2}, RoboVQA~\cite{sermanet2024robovqa}, and OpenEQA~\cite{majumdar2024openeqa}. These benchmarks assess multi-step planning in egocentric everyday scenarios, long-horizon reasoning for robotic manipulation tasks, and zero-shot understanding of embodied scenes in diverse environments, respectively. We observed that, \ours{} surpasses all comparison methods, including two proprietary models (i.e., GPT-4V~\cite{achiam2023gpt} and Gemini-2.5-Flash~\cite{comanici2025gemini}), exceeding the runner-up by 2.4\% on EgoPlan-Bench2, 5.5 BLEU score on RoboVQA, and 1.1 points on OpenEQA. These results demonstrate that \ours{} effectively handles complex planning sequences and extended reasoning horizons while generalizing to novel environments, showcasing robust capabilities for scene comprehension and multi-step task execution in embodied AI applications.

\begin{figure*}[t!]
    \centering
    \includegraphics[width=1.0\linewidth]{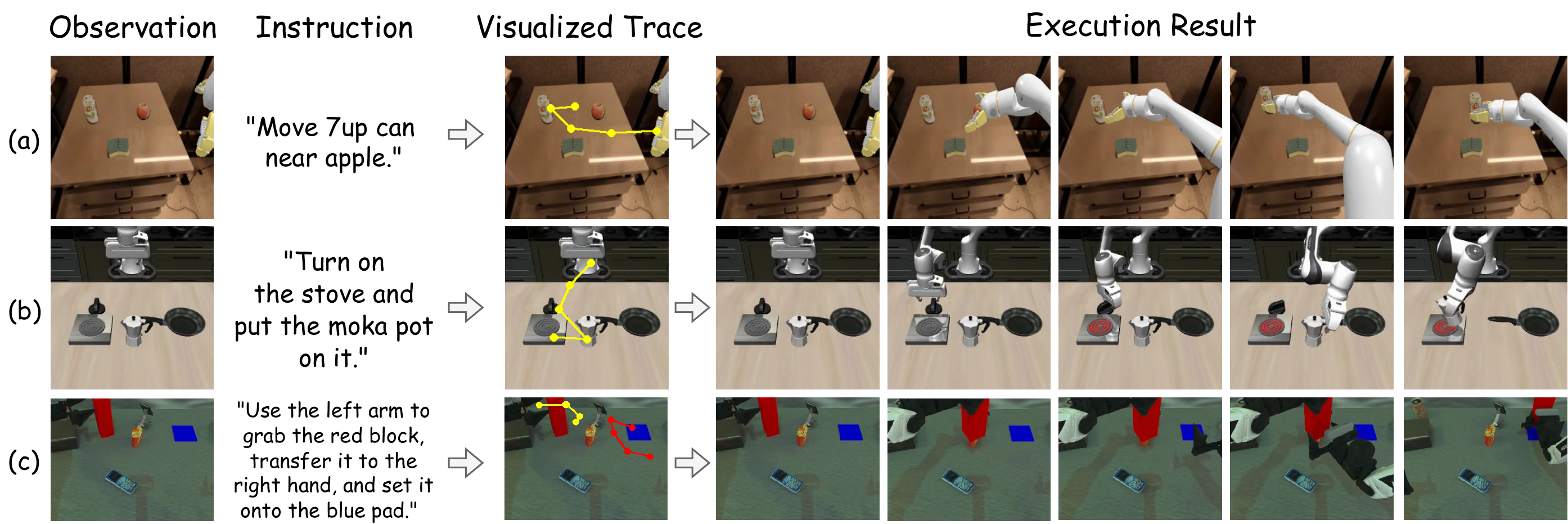}
    \caption{\textbf{Visualization of predicted visual trajectories and action execution results on long-horizon tasks.} Examples from (a) SimplerEnv-Google, (b) LIBERO-Long, and (c) RoboTwin2.0-Hard with long (278) steps. Yellow traces indicate single-arm/left gripper trajectories; red traces indicate right gripper trajectories for bimanual tasks.}
    \vspace{-2mm}
    \label{fig:visualize_manipulation}
\end{figure*}

\begin{figure*}[t!]
    \centering
    \includegraphics[width=1.0\linewidth]{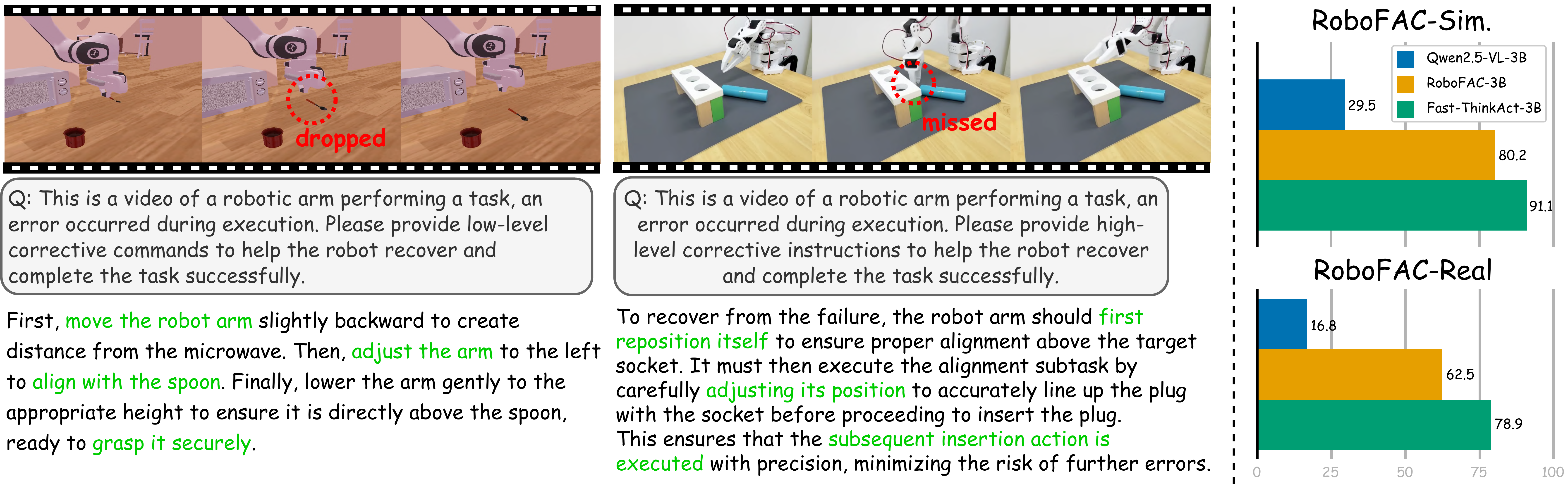}
    \vspace{-5mm}
    \caption{\textbf{Failure recovery capability on RoboFAC~\cite{lu2025robofac}.} Left: Qualitative examples (from both simulation and real robot) of corrective guidance for manipulation errors. Right: Quantitative evaluation on simulation (RoboFAC-Sim) and real-robot (RoboFAC-Real) settings.}
    \label{fig:visualize_robofac}
\end{figure*}

\subsection{Analysis of Fast-ThinkAct}

\paragraph{Reasoning Enables Long-Horizon Planning.}

\begin{wrapfigure}{R}{0.45\textwidth}
    \centering
    \vspace{-3mm}
    \includegraphics[width=\linewidth]{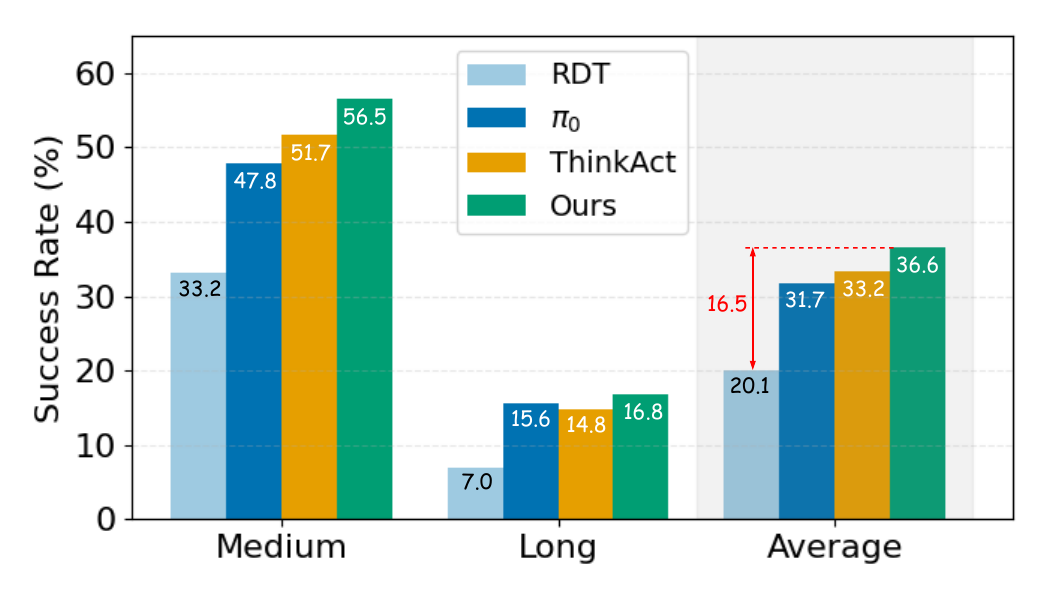}
    \caption{\textbf{Few-shot adaptation results on RoboTwin2.0 benchmark.} We use 10 demonstrations per task for fine-tuning.}
    \label{fig:exp_10shot}
\end{wrapfigure}

We analyze \ours{}'s capability on long-horizon tasks in Tab.~\ref{tab:robotwin} and Fig.~\ref{fig:visualize_manipulation}. We focus on long-horizon tasks (average length exceeding 270 steps) in RoboTwin2.0~\cite{chen2025robotwin2} that require multi-step reasoning and extended planning horizons. As shown in Tab.~\ref{tab:robotwin}, \ours{} achieves average scores of 48.8 and 16.8 on easy and hard settings of long-horizon tasks, respectively, surpassing RDT (35.0/12.3) and ThinkAct (42.8/15.3). Fig.~\ref{fig:visualize_manipulation} visualizes predicted 2D visual traces and execution results on representative tasks from SimplerEnv-Google~\cite{li24simpler}, LIBERO-Long~\cite{liu2023libero}, and RoboTwin2.0~\cite{chen2025robotwin2}. For example, the LIBERO-Long task requires sequentially turning on the stove and placing a moka pot on it, while the RoboTwin2.0 handover task requires bimanual coordination to transfer a block between grippers. The visual traces successfully predict feasible solution paths, with their corresponding representations serving as visual planning guidance for successful execution. These results demonstrate that our compact latent reasoning effectively supports long-horizon planning in complex manipulation scenarios.

\paragraph{Reasoning Enables Failure Recovery.}

\begin{wrapfigure}{R}{0.5\textwidth}
    \centering
    \includegraphics[width=1.0\linewidth]{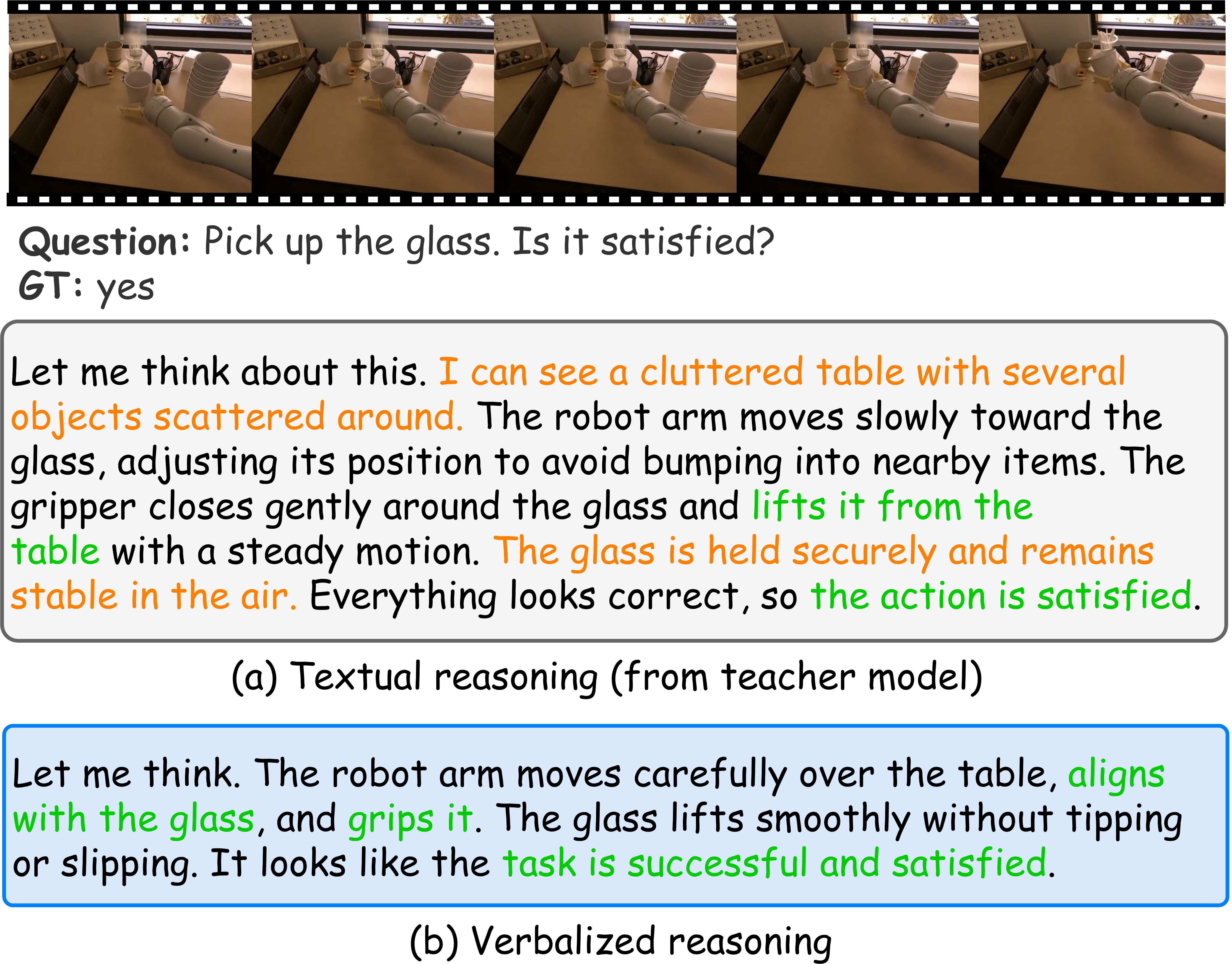}
    \caption{\textbf{Reasoning trace comparison on RoboVQA.} (a) Teacher's textual reasoning. (b) Student's verbalized latent reasoning. \textcolor{YellowGreen}{Green}: relevant content; \textcolor{orange}{orange}: less relevant content.}
    \label{fig:visualize_reasoning}
\end{wrapfigure}

A key advantage of reasoning-based VLAs~\cite{huang2025thinkact,abdolmaleki2025gemini} is their ability to identify runtime failures and provide corrective guidance for recovery. To evaluate this capability, we conduct experiments on RoboFAC~\cite{lu2025robofac}, a benchmark specifically designed to assess failure identification and correction in embodied VLMs. As shown in Fig.~\ref{fig:visualize_robofac}, \ours{} substantially outperforms the second-best baseline RoboFAC-3B~\cite{lu2025robofac} by 10.9 points on the simulation split and 16.4 points on the real-world split. The qualitative examples demonstrate \ours{}'s ability to reason over manipulation videos, identify failures, and propose recovery steps. For instance, in the right example where the target object drops mid-execution, \ours{} generates a concrete recovery plan: first moving the arm backward to create space, then adjusting laterally to align with the target object, and finally lowering to the appropriate height for a secure grasp. These results demonstrate that our latent reasoning supports both fast task execution and crucial failure analysis capabilities essential for robust robotic manipulation.

\paragraph{Reasoning Enables Few-Shot Adaptation.}

\begin{wraptable}{R}{0.55\textwidth}
    \centering
    \caption{\textbf{Ablation study of training objectives and learning stages.} Note that Fast-ThinkAct w/o $\mathcal{L}_{\text{verb}}, \mathcal{L}_{\text{distill}}$ denotes the student VLM $\mathcal{F}_\theta$ trained without the corresponding loss components.}
    \resizebox{\linewidth}{!}{
        \begin{tabular}{lcccc}
            \toprule
            \textbf{Method} & \textbf{EgoPlan} & \textbf{RoboVQA} & \textbf{OpenEQA} & \textbf{Average} \\
            \midrule
            \textbf{\ours{}} & \textbf{46.4} & \textbf{60.8} & \textbf{51.2} & \textbf{52.8} \\
            \midrule
            \; w/o $\mathcal{L}_\text{verb}$ & 42.1 & 53.8 & 49.5 & 48.5 \\
            \; w/o $\mathcal{L}_\text{verb},\mathcal{L}_\text{distill}$ & 41.6 & 52.7 & 48.9 & 47.7 \\
            \midrule
            Textual Teacher $\mathcal{F}_\theta^T$ & 41.7 & 58.2 & 49.4 & 49.8 \\
            SFT + CoT-SFT & 40.0 & 46.1 & 48.8 & 45.0 \\
            SFT only & 40.5 & 53.6 & 45.3 & 46.5 \\
            \bottomrule
        \end{tabular}
    }
    \label{tab:ablation_stage_loss}
\end{wraptable}

To assess how reasoning capability improves few-shot adaptation, we conduct few-shot experiments on the RoboTwin2.0 benchmark~\cite{chen2025robotwin2}, fine-tuning models using only 10 demonstrations per task. As illustrated in Fig.~\ref{fig:exp_10shot}, \ours{} significantly enhances our adopted action model RDT~\cite{liu2024rdt} and outperforms the state-of-the-art VLAs, including $\pi_0$~\cite{black2024pi_0} and ThinkAct~\cite{huang2025thinkact} on both medium and long-horizon tasks. Notably, our method achieves these gains while operating with significantly lower reasoning latency compared to ThinkAct, highlighting the advantage of efficient yet effective reasoning for few-shot action adaptation in complex robot manipulation scenarios.

\paragraph{Visualization of Verbalizable Latent Reasoning.}
In Fig.~\ref{fig:visualize_reasoning}, we compare the teacher's textual reasoning with the student's verbalized latent reasoning on RoboVQA. While both capture task-relevant information (\textcolor{YellowGreen}{green}), the teacher generates verbose outputs with less directly relevant content (\textcolor{orange}{orange}), whereas our student produces more concise and focused responses when verbalized. This demonstrates that our preference-guided distillation not only reduces computational cost but also distills concise reasoning patterns while filtering out redundant information.

\paragraph{Ablation Study.}

In Tab.~\ref{tab:ablation_stage_loss}, we ablate training stages and loss components. Starting from the full \ours{}, removing $\mathcal{L}_\text{verb}$ causes performance drops as latent CoTs lack preference-based guidance to align with high-quality reasoning and suppress low-quality patterns. Further removing $\mathcal{L}_\text{distill}$ leads to additional decline, indicating that aligning trajectory-level representations is crucial for transferring visual planning capabilities. Comparing training strategies, CoT-SFT underperforms SFT on EgoPlan-Bench2 and RoboVQA but improves on OpenEQA, suggesting naïve chain-of-thought supervision benefits open-ended QA but introduces verbosity that hinders structured reasoning tasks. Our preference-guided approach distills high-quality reasoning while maintaining efficiency. This validates the necessity of our proposed distillation framework and visual trajectory alignment. We provide additional ablation studies in the supplementary material.

\section{Conclusion}
\label{sec:conclusion}

We presented \textit{\ours{}}, an efficient reasoning framework for vision-language-action tasks that achieves compact yet expressive planning through verbalizable latent reasoning. By distilling lengthy textual reasoning into compact latent representations via preference-guided distillation and visual trajectory alignment, our approach bridges high-level embodied reasoning with low-level action execution through reasoning-enhanced policy learning. Extensive experiments across diverse robotic manipulation and embodied reasoning benchmarks demonstrate that \ours{} achieves strong performance with significantly reduced inference latency while enabling effective long-horizon planning, few-shot adaptation, and failure recovery capabilities.

\noindent\textbf{Limitations and Future Works.} As our verbalizer $\mathcal{V}_\psi$ is built upon a pre-trained LLM, it inevitably inherits language model limitations, including hallucination, occasionally producing plausible but inaccurate descriptions. However, this does not affect action execution during inference, as the verbalizer serves only for interpretability while action prediction uses the grounded latent representations from visual plan distillation. To further improve the faithfulness of verbalized reasoning, we can consider incorporating grounding-aware objectives or hallucination suppression techniques in future work.

\clearpage
\appendix

\newcommand{\refcolor}[1]{\textcolor{nvidiagreen}{#1}}

\section{Additional Experimental Setup}

\subsection{Algorithm}

\begin{algorithm}[H]
    \SetAlgoLined
    \caption{Training \ours{} (Sec.~\refcolor{3.2})}\label{algo:train}
    \KwIn{CoT-SFT checkpoint $\mathcal{F}_{\theta_0}$, training data $\mathcal{D}$, rollout size $N$, latent reasoning steps $M$, number of waypoints $K$, total iterations $T_{\text{total}}$}
    \KwOut{Trained student model $\mathcal{F}_\theta$}
    
    \tcp{Initialize models}
    $\mathcal{F}_\theta^T \leftarrow \mathcal{F}_{\theta_0}$, $\mathcal{F}_\theta \leftarrow \mathcal{F}_{\theta_0}$\;
    Initialize verbalizer $\mathcal{V}_\psi$ from pre-trained LLM\;
    $t \leftarrow 0$\;
    
    \While{$t < T_{\text{total}}$}{
        Sample batch $(o, l, \hat{p})$ from $\mathcal{D}$\;\tcp{Suppose bs=1 for simplicity}
        \vspace{0.5em}
        
        \tcp{Teacher GRPO training}
        Generate $N$ rollouts $\{\tau_i\}_{i=1}^N$ from $\mathcal{F}_\theta^T(o, l)$\;
        Compute trajectory rewards $\{r_i\}_{i=1}^N$\;
        Compute group-wise advantages $\{A_i\}_{i=1}^N$\;
        Update $\mathcal{F}_\theta^T$ with $\mathcal{J}_{\text{GRPO}}$ (Eq.~\refcolor{1})\;
        $\tau^+ \leftarrow \arg\max_i A_i$, $\tau^- \leftarrow \arg\min_i A_i$ (Eq.~\refcolor{3}) \tcp*{For student distillation}
        $h_t^T \leftarrow$ hidden state of $\tau^+$ at \texttt{<answer>} token from $\mathcal{F}_\theta^T$ \tcp*{For distillation loss}

        \vspace{0.5em}
        \tcp{Student latent distillation}
        $\mathbf{z} = \{z_m\}_{m=1}^M \leftarrow \mathcal{F}_\theta(o, l)$ \tcp*{Perform auto-regressive latent reasoning}
        Compute $\mathcal{L}_{\text{verb}}$ with $\mathbf{z}$, $\mathcal{V}_\psi$, $\tau^+$, $\tau^-$ (Eq.~\refcolor{4})\;
        
        Forward $K$ spatial tokens from $\mathcal{F}_\theta(o, l, \mathbf{z})$ to obtain $h_t$ and $\{h^\prime(\mathbf{s}_i)\}_{i=1}^K$\;
        Compute $\mathcal{L}_{\text{distill}}$ with $h_t^T$, $h_t$ (Eq.~\refcolor{5})\;
        Compute $\mathcal{L}_{\text{ans}}$ with $\{h^\prime(\mathbf{s}_i)\}_{i=1}^K$, $\hat{p}$ (Eq.~\refcolor{6})\;
        
        Update $\mathcal{F}_\theta$ with $\mathcal{L}_{\text{student}} = \mathcal{L}_{\text{verb}} + \mathcal{L}_{\text{distill}} + \mathcal{L}_{\text{ans}}$\;
        
        $t \leftarrow t + 1$\;
    }
    
    \Return{$\mathcal{F}_\theta$}\;
\end{algorithm}

Algorithm~\ref{algo:train} presents the complete training procedure corresponding to Sec.~\refcolor{3.2}. It shows how we jointly optimize the teacher model with GRPO and distill its reasoning into the student's compact latent representations.

\subsection{Implementation Details}

Our implementation follows the setup described in Sec.~\refcolor{4.1} of the main paper. Here we provide additional details. The verbalizer $\mathcal{V}_\psi$ is initialized from a small LLM, Qwen3-0.6B, with cross-attention layers inserted at each layer to condition on latent CoTs $\mathbf{z}$. For the student model training, in the first 3,000 iterations, we replace verbalization loss $\mathcal{L}_{\text{verb}}$ with language modeling loss using $\tau^+$ as ground truth to warm up $\mathcal{V}_\psi$'s alignment with the latent representations $\mathbf{z}$. We then freeze $\mathcal{V}_\psi$ and use the $\mathcal{L}_{\text{verb}}$ for the remaining 1,500 iterations. The student $\mathcal{F}_\theta$ is optimized throughout both phases. For waypoint prediction in Eq.~\refcolor{6}, each $p_i \in \mathbb{R}^6$ encodes coordinates in the format $[x_{\text{single}}, y_{\text{single}}, x_{\text{left}}, y_{\text{left}}, x_{\text{right}}, y_{\text{right}}]$, where the first two dimensions are for single-arm and the last four are for bimanual robot. For ground-truth $\hat{p}i$, we fill the corresponding dimensions based on robot type and mask out the unused dimensions when computing $\mathcal{L}_{\text{ans}}$. For GRPO training, we follow the configuration of ThinkAct~\cite{huang2025thinkact}, using rollout size $N=5$. Following~\cite{lee2025molmoact}, we set the number of waypoints in trajectory to $K=5$. We use $M=6$ latent reasoning tokens, with ablation study provided in Fig.~\ref{fig:supp:ablate_steps}.

During reasoning-enhanced policy learning, for SimplerEnv~\cite{li24simpler} evaluation, to ensure fair comparison with previous works~\cite{kim24openvla, lee2025molmoact}, we initialize $\pi_\phi$ from DiT-Policy~\cite{chi2023diffusion} pre-trained on the same OXE dataset~\cite{o2024open, kim24openvla} and conduct reasoning-enhanced policy learning (Sec.~\refcolor{3.3}) using the same OXE data. For LIBERO~\cite{liu2023libero} and RoboTwin2.0~\cite{chen2025robotwin2} evaluations, we initialize $\pi_\phi$ from RDT~\cite{liu2024rdt}, which has demonstrated strong performance on RoboTwin2.0, and conduct policy learning using OXE~\cite{o2024open} and static ALOHA datasets~\cite{shi2023waypoint, zhao2023learning}. Our method further enhances RDT's manipulation capabilities on both benchmarks. The use of different action models also demonstrates that our approach is agnostic to the underlying action model choice.

\subsection{Training Data Details}

\subsubsection{Dataset Sources}

\paragraph{2D Visual Trace of Manipulation Tasks.} For single-arm manipulation, we utilize 2D visual trajectories labeled by MolmoAct~\cite{lee2025molmoact} from the Open X-Embodiment (OXE) dataset~\cite{o2024open}, comprising approximately 1.3M trajectories. For bimanual manipulation, we extract dual-arm visual trajectories from the AIST dataset~\cite{aist2025aist}, resulting in approximately 92K trajectory samples. Specifically, we first use Molmo-72B~\cite{deitke2024molmo} to detect left and right gripper positions (following~\cite{lee2025molmoact}) in the first frame, then apply CoTracker3~\cite{karaev2025cotracker3} to track and parse the manipulation trajectories throughout the video sequences.

\paragraph{RoboFAC~\cite{lu2025robofac}.} RoboFAC is a robotic failure analysis dataset containing 9,440 erroneous manipulation trajectories across 16 tasks in both simulated and real-world environments. We utilize the training set with 64K QA pairs covering various failure types for developing failure identification and correction planning capabilities.

\paragraph{RoboVQA~\cite{sermanet2024robovqa}.} RoboVQA contains robot manipulation videos with QA tasks covering task understanding. The dataset includes approximately 5K long-horizon and 92K medium-horizon video sequences from diverse robotic platforms, resulting in total 798K QA pairs. Videos are annotated with multiple questions probing spatial reasoning, action prediction, and task comprehension.

\paragraph{ShareRobot~\cite{ji2025robobrain}.} ShareRobot is a large-scale dataset collected by RoboBrain~\cite{ji2025robobrain}, containing over 1M QA pairs covering task planning, object affordances, and manipulation strategies across diverse robot embodiments and scenes. The dataset features fine-grained annotations linking task descriptions to frame-level execution details, facilitating learning of transferable manipulation knowledge.

\paragraph{EgoPlan-Bench~\cite{chen2023egoplan}.} EgoPlan-Bench features egocentric videos of daily activities annotated with task planning information including goals, execution history, and current states. The dataset contains approximately 53K video-text pairs for training long-horizon planning and progress tracking capabilities from egocentric view.

\paragraph{Video-R1-CoT~\cite{feng2025video}.} Video-R1 comprises 165K video question-answer pairs with chain-of-thought reasoning annotations generated by large-scale vision-language models. The dataset covers diverse reasoning domains including mathematical logic, spatial understanding, OCR, and visual analytics. All samples are quality-filtered to ensure annotation consistency and correctness.

\paragraph{PixMo~\cite{deitke2024molmo}.} PixMo is a general-purpose vision-language dataset with diverse image captions and question-answer pairs. Following MolmoAct~\cite{lee2025molmoact}, we incorporate PixMo dataset to preserve general visual understanding and prevent catastrophic forgetting when training on embodied dataset. Specifically, we use approximately 726K samples from the \texttt{ask\_model\_anything}, \texttt{cap}, and \texttt{cap-qa} splits.

\subsubsection{Data Processing and Formatting}

\paragraph{Supervised Fine-Tuning (SFT).} To enhance foundational embodied knowledge, we perform supervised fine-tuning on approximately 4M samples combining 2D visual trajectories from MolmoAct~\cite{lee2025molmoact} and AIST~\cite{aist2025aist}, along with QA data from PixMo~\cite{deitke2024molmo}, RoboFAC~\cite{lu2025robofac}, RoboVQA~\cite{sermanet2024robovqa}, ShareRobot~\cite{ji2025robobrain}, and EgoPlan~\cite{chen2023egoplan}. This stage enables the model to acquire basic visual understanding, task comprehension, and manipulation knowledge across diverse embodiments and scenarios.

\paragraph{Chain-of-Thought SFT (CoT-SFT).} To develop reasoning capabilities while preserving embodied understanding, we sample 5\% from the SFT data (approximately 200K samples) and augment with 165K samples from Video-R1-CoT~\cite{feng2025video}. For data with CoT annotations, we format prompts to elicit structured reasoning enclosed in \texttt{<think>} tags followed by answers in \texttt{<answer>} tags; for data without CoT annotations, we prompt for direct answers only. This enables the model to learn reasoning capabilities from CoT-annotated data and generalize them to embodied tasks.

\paragraph{Teacher-Student Training.} Building upon the CoT-SFT checkpoint, we curate a balanced training set by sampling approximately 5,000 instances from each dataset and data type, totaling nearly 50K samples. We adopt the prompt formatting strategy from CoT-SFT for both teacher GRPO training and student latent distillation. We train both the teacher with GRPO and the student with latent distillation (as detailed in Sec.~\refcolor{3.2}) on this data, efficiently transferring high-quality reasoning patterns into compact latent representations.

\begin{table*}[t]
\centering
\caption{\textbf{Quantitative results with larger model size (7B or 8B) on embodied reasoning benchmarks.}}
\resizebox{1.0\textwidth}{!}{
\small
\begin{tabular}{lccccccccccccc}
\toprule
\textbf{Method} & \multicolumn{5}{c}{\textbf{EgoPlan-Bench2}} & \multicolumn{5}{c}{\textbf{RoboVQA}} & \textbf{OpenEQA} & \cellcolor{gray!15}\textbf{Overall} \\
\cmidrule(lr){2-6} \cmidrule(lr){7-11}
 & Daily. & Work. & Rec. & Hobbies & Avg. & B-1 & B-2 & B-3 & B-4 & B-Avg. & Score & \cellcolor{gray!15}Avg. \\
\midrule
InternVL2.5-8B~\cite{chen2024expanding} & 36.2 & 28.7 & 34.4 & 35.4 & 33.5 & 40.5 & 33.3 & 29.6 & 27.5 & 32.7 & 54.4 & \cellcolor{gray!15}40.2 \\
InternVL3-8B~\cite{zhu2025internvl3} & 38.5 & 32.9 & 36.1 & 37.2 & 36.2 & 44.3 & 36.5 & 31.6 & 28.9 & 35.3 & 55.5 & \cellcolor{gray!15}42.3 \\
NVILA-8B~\cite{liu2024nvila} & 35.8 & 28.7 & 37.2 & 35.4 & 33.7 & 42.7 & 39.7 & 37.6 & 36.1 & 39.0 & 54.0 & \cellcolor{gray!15}42.2 \\
Qwen2.5-VL-7B~\cite{bai2025qwen2} & 31.4 & 26.7 & 29.5 & 28.6 & 29.1 & 47.8 & 41.2 & 36.2 & 33.7 & 39.7 & 50.8 & \cellcolor{gray!15}39.9 \\
Magma-8B~\cite{yang2025magma} & 32.1 & 25.7 & 34.4 & 29.3 & 29.8 & 38.6 & 31.5 & 28.1 & 26.7 & 31.2 & 49.1 & \cellcolor{gray!15}36.7 \\
RoboBrain2.0-7B~\cite{team2025robobrain2} & 39.4 & 27.0 & 33.9 & 32.2 & 33.2 & 44.9 & 38.2 & 34.7 & 33.5 & 37.8 & 51.1 & \cellcolor{gray!15}40.7 \\
ThinkAct-7B~\cite{huang2025thinkact} & 50.1 & 49.8 & 44.8 & 45.2 & 48.2 & 69.1 & 61.8 & 56.0 & 52.4 & 59.8 & 56.2 & \cellcolor{gray!15}54.7 \\
\midrule
\ours{}-7B & 51.3 & 47.3 & 41.5 & 45.9 & 47.5 & 70.4 & 63.3 & 57.3 & 53.2 & 61.1 & 59.0 & \cellcolor{gray!15}\textbf{55.9} \\
\bottomrule
\end{tabular}
}
\label{tab:supp_er_7b}
\vspace{-2mm}
\end{table*}

\begin{table}[t!]
    \centering
    \caption{\textbf{Results on LIBERO and SimplerEnv benchmarks with additional ThinkAct-3B comparison.}}
    \resizebox{0.65\columnwidth}{!}{
        \small
        \begin{tabular}{lccc}
            \toprule
            \textbf{Method} & \textbf{LIBERO} & \textbf{SimplerEnv-Google} & \textbf{Latency ($\downarrow$)} \\
            \midrule
            OpenVLA-7B~\cite{kim24openvla} & 76.5 & 40.2 & N/A \\
            CoT-VLA-7B~\cite{zhao2025cot} & 83.9 & N/A & N/A \\
            ThinkAct-7B~\cite{huang2025thinkact} & 84.4 & 68.3 & 7513 \\
            MolmoAct-7B~\cite{lee2025molmoact} & 86.8 & 64.9 & 6723 \\
            \midrule
            ThinkAct-3B~\cite{huang2025thinkact} & 83.1 & 64.7 & 5674 \\
            \textbf{Fast-ThinkAct-3B} & \textbf{89.7} & \textbf{68.7} & \textbf{805 (\textcolor{YellowGreen}{$\downarrow$7.0$\times$})} \\
            \bottomrule
        \end{tabular}
    }
    \label{tab:supp_libero_simpler}
\end{table}
\begin{table}[t!]
    \centering
    \caption{\textbf{Comparison with efficient textual reasoning methods.}}
    \resizebox{0.75\linewidth}{!}{
        \begin{tabular}{lcccc}
            \toprule
            \makecell{\textbf{Method}} &
            \makecell{\textbf{EgoPlan-}\\\textbf{Bench2}} &
            \makecell{\textbf{RoboVQA}} &
            \makecell{\textbf{OpenEQA}} &
            \makecell{\textbf{Average}} \\
            \midrule
            Textual Teacher $\mathcal{F}_\theta^T$ & 41.7 & 58.2 & 49.4 & 49.8 \\
            $\mathcal{F}_\theta^T$ Inference w/o thinking & 42.7 & 55.0 & 41.7 & 46.5 \\
            $\mathcal{F}_\theta^T$ Inference w/ 6 textual tokens & 39.3 & 53.0 & 46.5 & 46.3 \\
            $\mathcal{F}_\theta^T$ w/ RL Length-Penalty~\cite{arora2025training} & 41.2 & 57.5 & 44.7 & 47.8 \\
            \midrule
            Fast-ThinkAct-3B & \textbf{46.4} & \textbf{60.8} & \textbf{52.8} & \textbf{53.3} \\
            \bottomrule
        \end{tabular}
    }
    \label{tab:supp:compare_efficient}
\end{table}

\subsection{Evaluation Setup}

\subsubsection{Embodied Reasoning Benchmarks}

We evaluate on three benchmarks assessing different aspects of embodied reasoning. EgoPlan-Bench2~\cite{chen2023egoplan} tests egocentric task planning across 24 daily-life scenarios with 1,321 multiple-choice questions, measuring accuracy in predicting next steps given task goals and progress history. RoboVQA~\cite{sermanet2024robovqa} evaluates visual reasoning in manipulation contexts through 1,893 free-form QA pairs from robot and human demonstrations, assessed via BLEU score. OpenEQA~\cite{majumdar2024openeqa} assesses spatial and functional understanding through 1,600+ questions spanning 180+ real-world environments, evaluated using LLM-based scoring aligned with human preferences. These benchmarks comprehensively evaluate embodied reasoning capability across planning, manipulation, and spatial understanding.

\subsubsection{Robotic Manipulation Benchmarks}

\begin{figure*}[t!]
    \centering
    \includegraphics[width=0.96\linewidth]{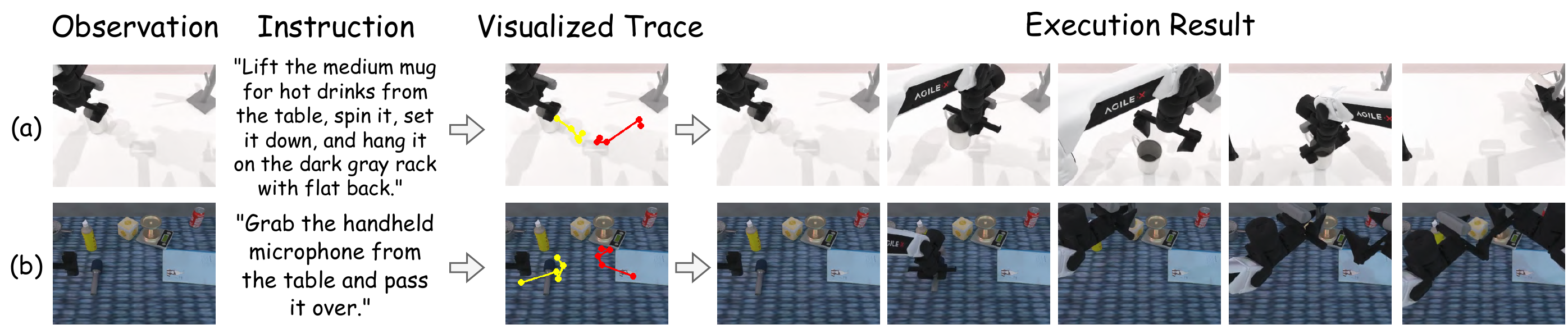}
    \caption{\textbf{Visualization of predicted visual trajectories and action execution results on RoboTwin2.0.} Yellow traces indicate left gripper trajectories; red traces indicate right gripper trajectories for bimanual tasks.}
    \vspace{-2mm}
    \label{fig:supp:visualize_manipulation}
\end{figure*}

\begin{figure*}[t!]
    \centering
    \includegraphics[width=0.9\linewidth]{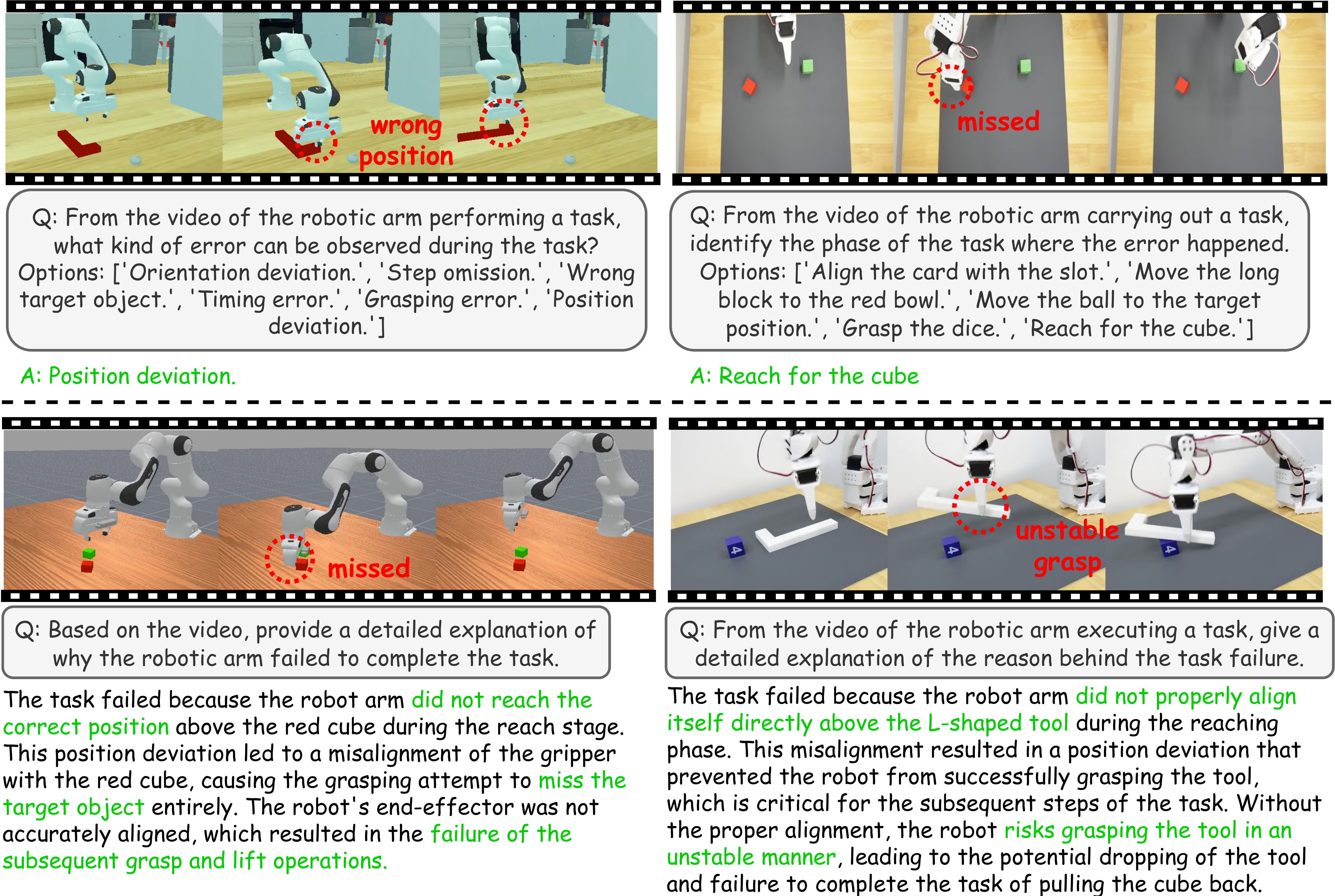}
    \caption{\textbf{Failure identification and analysis capabilities on RoboFAC~\cite{lu2025robofac}.} Top row shows identification of failure types and execution stages. Bottom row demonstrates failure root cause analysis.}
    \label{fig:supp:visualize_robofac}
\end{figure*}

\begin{figure}[t!]
    \centering
    \includegraphics[width=0.60\linewidth]{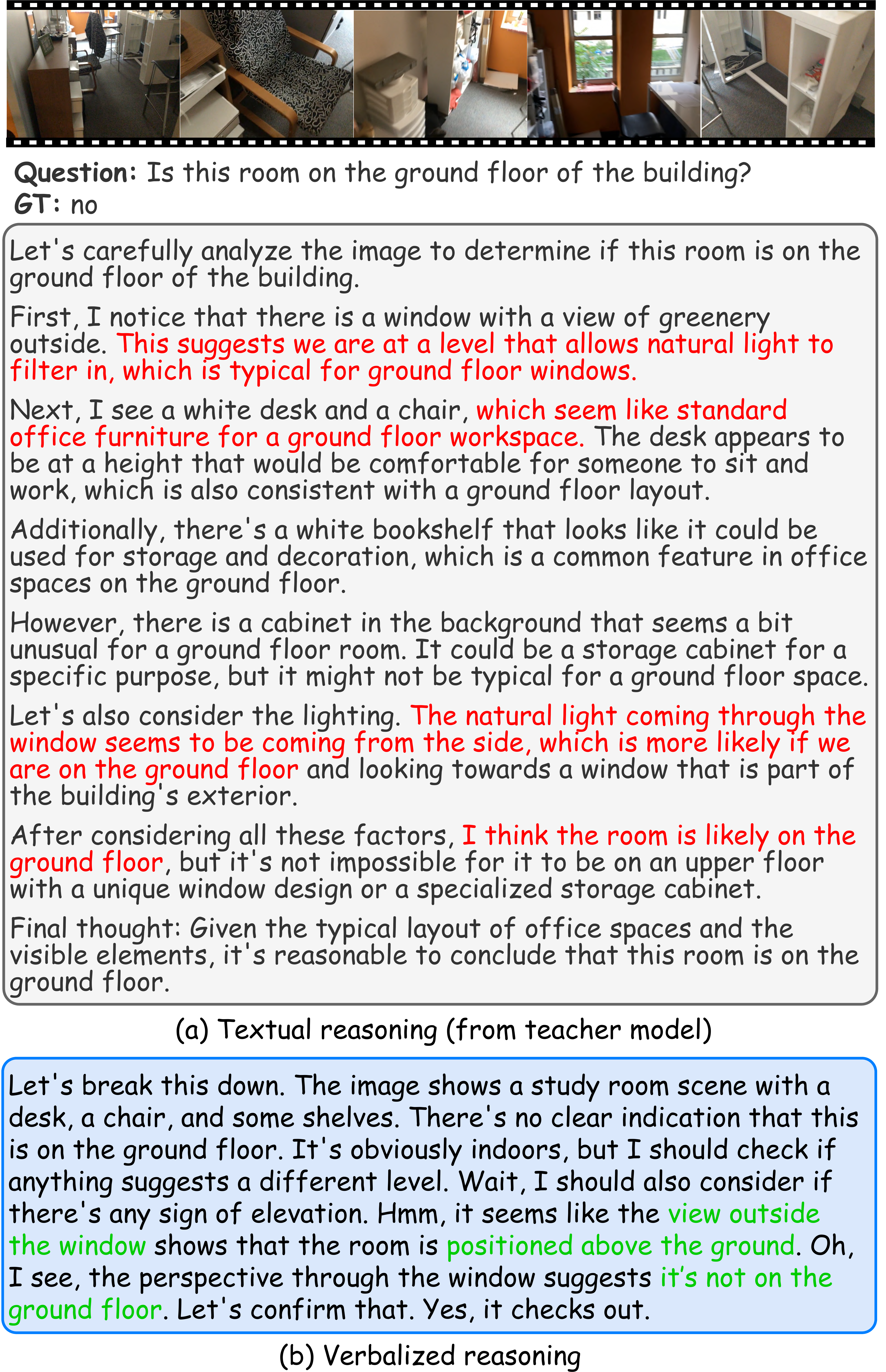}
    \caption{\textbf{Reasoning trace comparison on OpenEQA.} (a) Teacher's textual reasoning. (b) Student's verbalized latent reasoning. \textcolor{YellowGreen}{Green}: reasonable reasoning trace; \textcolor{red}{red}: incorrect trace.}
    \label{fig:supp:visualize_reasoning}
\end{figure}

We evaluate on three simulation benchmarks covering diverse manipulation scenarios. SimplerEnv~\cite{li24simpler} provides manipulation tasks with strong sim-to-real correlation, featuring diverse visual variations in lighting, textures, backgrounds, and camera poses. Following MolmoAct~\cite{lee2025molmoact}, we evaluate on the Google Robot tasks using the standard protocol~\cite{kim24openvla, lee2025molmoact} of directly evaluating on SimplerEnv after training on OXE. LIBERO~\cite{liu2023libero} targets different generalization challenges through four task suites: spatial layout variation (LIBERO-Spatial), object diversity (LIBERO-Object), goal variation (LIBERO-Goal), and long-horizon planning with mixed variations (LIBERO-Long). We evaluate each suite over 500 trials using 3 random seeds following prior works~\cite{kim24openvla, lee2025molmoact}. RoboTwin2.0~\cite{chen2025robotwin2} features challenging bimanual manipulation with easy and hard difficulty settings, where the hard setting introduces domain randomization including clutter, lighting variations, diverse textures, and height changes. Following the original protocol, we train on 50 clean expert demonstrations per task and evaluate with 100 rollouts under both settings. We assess 10 tasks categorized into short, medium, and long horizons based on demonstration lengths.

\section{Additional Experiment Results}

\subsection{Additional Quantitative Results}


\paragraph{Results of Larger Model Size.} To demonstrate the scalability of our approach, we apply Fast-ThinkAct to a larger backbone, Qwen2.5-VL-7B, and evaluate its performance on embodied reasoning benchmarks. As shown in Tab.~\ref{tab:supp_er_7b}, Fast-ThinkAct consistently achieves strong performance across EgoPlan-Bench2~\cite{qiu2024egoplan2}, RoboVQA~\cite{sermanet2024robovqa}, and OpenEQA~\cite{majumdar2024openeqa}, validating that our latent reasoning distillation method effectively scales to larger model backbones.

\paragraph{Performance Comparison with ThinkAct-3B.} Tab.~\ref{tab:supp_libero_simpler} presents detailed numerical results corresponding to Fig.~\refcolor{3} with additional ThinkAct-3B results. At the same 3B model size, Fast-ThinkAct achieves notable performance gains (89.7 vs. 83.1 on LIBERO, 68.7 vs. 64.7 on SimplerEnv-Google) while dramatically improving efficiency with \textbf{\textcolor{YellowGreen}{7$\times$}} faster inference (805ms vs. 5674ms).

\paragraph{Comparison with Efficient Reasoning Baselines.} Table~\ref{tab:supp:compare_efficient} compares our method with efficient textual reasoning alternatives applied to the textual teacher $\mathcal{F}_\theta^T$. We evaluate three baselines: removing reasoning during inference entirely (0 tokens), constraining the teacher to generate only 6 textual tokens during inference, and applying RL training with a length penalty~\cite{arora2025training} to encourage concise reasoning ($\sim$50 tokens). These achieve 46.5, 46.3, and 47.8 respectively, all degrading from the teacher's 49.8. In contrast, \ours{} uses only 6 latent tokens and achieves \textbf{53.3}, demonstrating superior efficiency and performance.

\subsection{Additional Qualitative Results}

\paragraph{Qualitative Robot Execution.} We provide qualitative robot execution comparisons between the base action model RDT~\cite{liu2024rdt} and \ours{} in the supplementary video \texttt{Fast-ThinkAct.mp4}. Our method shows substantial improvements on challenging robotic execution tasks, where reasoning capabilities provide better spatial understanding and coordination for successful manipulation.

\paragraph{Bimanual Manipulation Results.} In Fig.~\ref{fig:supp:visualize_manipulation}, we present visualized trajectories and execution results for \texttt{hanging mug} and \texttt{handover mic} tasks under easy and hard settings in RoboTwin2.0~\cite{chen2025robotwin2}. The hard setting includes different backgrounds and distractor objects. These examples show successful bimanual coordination where predicted waypoints accurately guide both grippers through the manipulation sequence, demonstrating Fast-ThinkAct's spatial reasoning ability across varied visual conditions.

\paragraph{Failure Identification and Recovery.} In Fig.~\ref{fig:supp:visualize_robofac}, we demonstrate Fast-ThinkAct's failure identification and analysis capabilities, complementing the recovery planning shown in the main paper. The top row shows that Fast-ThinkAct identifies failure types (e.g., position deviation) and execution stages (e.g., reaching for the cube). The bottom row illustrates root cause analysis, for instance, in the bottom-right example, the model correctly infers that the failure to push the cube with an L-shaped tool stems from an improper initial grasp. These results demonstrate Fast-ThinkAct's comprehensive understanding of manipulation failures beyond recovery planning.

\paragraph{Verbalized Latent Reasoning.} Fig.~\ref{fig:supp:visualize_reasoning} visualizes teacher textual reasoning and student verbalized reasoning. While the student generates compact and correct (\textcolor{YellowGreen}{green}) reasoning, the teacher's lengthy output sometimes contains erroneous steps (\textcolor{red}{red}) that might degrade the performance.

\subsection{Additional Ablation Study and Analysis}

\paragraph{Additional Ablation Results on Manipulation Benchmarks.} Table~\ref{tab:supp:ablation_stage_loss} shows ablation results on LIBERO~\cite{liu2023libero}, SimplerEnv-Google~\cite{li24simpler}, and RoboTwin2.0~\cite{chen2025robotwin2}. Removing $\mathcal{L}_{\text{verb}}$ or $\mathcal{L}_{\text{distill}}$ progressively degrades performance, confirming their contributions. Our full model consistently outperforms the textual teacher and models without teacher-student training (CoT-SFT, SFT only), demonstrating the benefits of compact latent reasoning distillation.

\begin{table}[t!]
    \begin{minipage}{0.5\columnwidth}
        
        \centering
        \caption{\textbf{Additional ablation study of training objectives and learning stages on robot manipulation benchmarks.}}
        \label{tab:supp:ablation_stage_loss}
    
        \centering
        \resizebox{\linewidth}{!}{
        \begin{tabular}{lcccc}
            \toprule
            \makecell{\textbf{Method}} &
            \makecell{\textbf{LIBERO}} &
            \makecell{\textbf{SimplerEnv}\\\textbf{Google}} &
            \makecell{\textbf{RoboTwin2.0}} &
            \makecell{\textbf{Average}} \\
            \midrule
            \textbf{Fast-ThinkAct} & \textbf{89.7} & \textbf{68.7} & \textbf{46.1} & \textbf{68.2} \\
            \midrule
            w/o $\mathcal{L}_\text{verb}$ & 88.6 & 67.3 & 44.9 & 66.9 \\
            w/o $\mathcal{L}_\text{verb},\mathcal{L}_\text{distill}$ & 86.3 & 65.7 & 42.6 & 64.9 \\
            \midrule
            Textual Teacher & 88.5 & 67.3 & 45.8 & 67.2 \\
            SFT + CoT-SFT & 87.2 & 65.8 & 43.3 & 65.4 \\
            SFT only & 86.9 & 64.5 & 42.8 & 64.7 \\
            \bottomrule
        \end{tabular}
    }

    \end{minipage}\hfill
    \begin{minipage}{0.46\columnwidth}

        \centering
        \includegraphics[width=\linewidth]{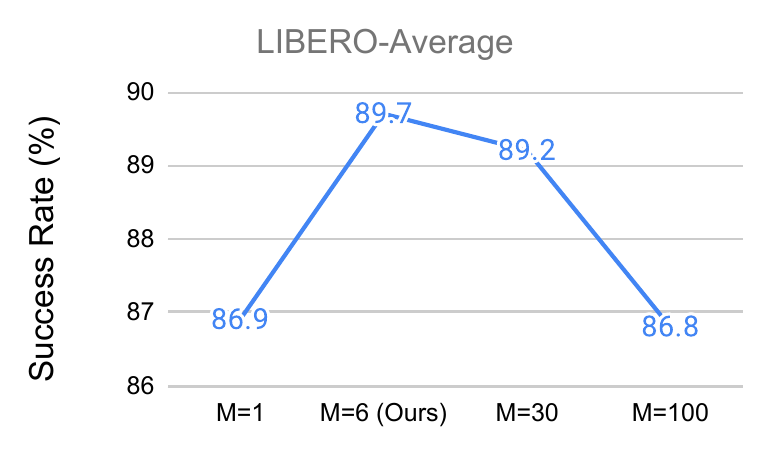}

        \caption{\textbf{Ablation of Latent Reasoning Steps $M$.}}
    \label{fig:supp:ablate_steps}
    \end{minipage}

\end{table}

\paragraph{Ablation Study on Action Model Conditioning.} In Sec.~\refcolor{3.3}, we extract visual latent planning $c_t$ from early-layer KV cache of spatial tokens to condition the action model. We compare this against using late-layer KV cache (last $N$ layers, where $N$ is the action model depth) and directly using spatial tokens' output hidden states. Our approach achieves \textbf{89.7} on LIBERO, outperforming late-layer KV at 88.3 and output hidden states at 87.1, demonstrating that early-layer representations better capture visual planning information for action prediction. Therefore, we adopt early-layer KV conditioning as our default configuration.

\paragraph{Ablation Study on Latent Reasoning Steps.} In Fig.~\ref{fig:supp:ablate_steps}, we study the effect of latent reasoning steps $M$. We observe that too few steps ($M=1$) limit reasoning capacity, while excessive steps ($M=30, 100$) might introduce redundant or noisy information. Therefore, we adopt $M=6$, which achieves optimal performance, as our default.

\clearpage
\setcitestyle{numbers}
\bibliographystyle{plainnat}
\bibliography{main}

\end{document}